\relax
\documentclass[letterpaper]{article} % DO NOT CHANGE THIS
\pdfoutput=1
\usepackage{aaai21}  % DO NOT CHANGE THIS
\usepackage{times}  % DO NOT CHANGE THIS
\usepackage{helvet} % DO NOT CHANGE THIS
\usepackage{courier}  % DO NOT CHANGE THIS
\usepackage[hyphens]{url}  % DO NOT CHANGE THIS
\usepackage{graphicx} % DO NOT CHANGE THIS
\usepackage{multirow}
\usepackage{latexsym}
\usepackage{microtype}
\usepackage{amsmath}
\usepackage{verbatim}
\usepackage{bm}
\usepackage{amsfonts}
\usepackage{graphicx}  % DO NOT CHANGE THIS
\usepackage{natbib}  % DO NOT CHANGE THIS AND DO NOT ADD ANY OPTIONS TO IT
\usepackage{caption} % DO NOT CHANGE THIS AND DO NOT ADD ANY OPTIONS TO IT
\title{Hierarchical Ranking for Answer Selection}

\author{

    Hang Gao,
    Mengting Hu,
    Renhong Cheng,
    Tiegang Gao\\
}
\affiliations{
    Nankai University\\
    knimet@mail.nankai.edu.cn

}

\begin{document}

\maketitle

\begin{abstract}
Answer selection is a task to choose the positive answers from a pool of candidate answers for a given question. In this paper, we propose a novel strategy for answer selection, called hierarchical ranking. We introduce three levels of ranking: point-level ranking, pair-level ranking, and list-level ranking. They formulate their optimization objectives by employing supervisory information from different perspectives to achieve the same goal of ranking candidate answers. Therefore, the three levels of ranking are related and they can promote each other. We take the well-performed compare-aggregate model as the backbone and explore three schemes to implement the idea of applying the hierarchical rankings jointly: the scheme under the Multi-Task Learning (MTL) strategy, the Ranking Integration (RI) scheme, and the Progressive Ranking Integration (PRI) scheme. Experimental results on two public datasets, WikiQA and TREC-QA, demonstrate that the proposed hierarchical ranking is effective. Our method achieves state-of-the-art (non-BERT) performance on both TREC-QA and WikiQA. 
\end{abstract}

\section{Introduction}

Answer selection is a basic task in question answering. Given a question and a list of candidate sentences, the machine needs to select the positive answers, which are sentences that can answer the question. Other sentences are called negative answers.

Recently, attention-based neural networks perform well in this task. A number of recent works \cite{he2015multi-perspective,santos2016attentive,bian2017a,wang2017bilateral} have proposed multiple variants of attention mechanisms for different scenarios. Besides, we observed that previous models mainly have three different usages of the supervisory signal.

\begin{figure}[t]
\centering
\includegraphics[width=0.45\textwidth]{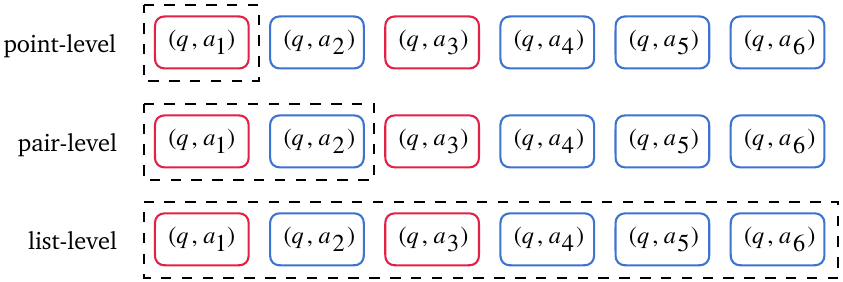}
\caption{Difference between the three levels of ranking. The dashed box shows the different granularities of different levels of ranking. The red box indicates a positive question-answer pair and the blue box indicates a negative one.}
\label{threelevelranking}
\end{figure}

Some works \cite{devlin2018bert:,tay2017learning,wang2017bilateral,shen2017inter-weighted,deng2019multi-task,shen2018knowledge-aware} regard the supervisory signals as the correlation between the question and answer. They utilize supervisory signals to learn to rank on point-level. This is intuitive and has been widely adopted.  Some works \cite{rao2016noise-contrastive,he2015multi-perspective,santos2016attentive} construct positive-negative pairs according to supervisory signals, and distinguish the positive answers and negative answers by learning their differences. This aims to rank the candidate answers on pair-level. Some other models \cite{bian2017a,wang2016a} treat the supervisory signals as the list structure information and learn to rank the candidate answers. This can be called learning to rank on list-level. In conclusion, they can learn to rank based on different levels of granularity.

Based on the experience of deep learning, different use methods of supervisory information have different optimization objectives in training. The point-level ranking focus on optimizing the relevance score of each question-answer pair. The pair-level ranking learns the correlation among candidates in a contrastive way, which distinguishes the positive and negative answers. The models are optimized by triplet ranking loss to allocate a higher score for positive pair $(Q,A^{+})$ than the negative one $(Q,A^{-})$. The list-level ranking aims to fit the prediction of a list of question-answer pairs with the ground-truth. In other words, they can rank the answer sentences through different optimization objectives, which shows that they are related and complementary. We argue that applying these hierarchical rankings jointly can bring performance improvement.

Under the idea of combining the hierarchical rankings, an intuitive solution is employing multi-task learning which treats the three rankings as three tasks. However, there exist internal relationships between the three optimization goals. For the same data, they focus on different levels of data granularity. This allows them to promote and inspire each other. For example, matching $(Q,A)$ pair better might improve the pair-level comparison and the list-level ranking might benefit from the pair-level approach. Simple multi-task learning might be insufficient to make full use of these relationships. 

In this paper, we propose a novel strategy for answer selection, called hierarchical ranking. Different from previous works that learn to rank at a fixed level, there are three levels of ranking in our strategy: point-level ranking, pair-level ranking, and list-level ranking. We design three schemes to implement the proposed hierarchical ranking strategy. Specifically, we first introduce the scheme under Multi-Task Learning (MTL) strategy. Then, we propose the Ranking Integration (RI) scheme to make better use of the internal relationships between the three levels of ranking. Finally, we explore the Progressive Ranking Integration (PRI) scheme, which further enhances the combination of internal relations.

In summary, the main contributions of this paper are as follows:
\begin{itemize}
    \item We revisit the three classical methods for answer selection. We propose a novel strategy for exploiting their merits and internal  relationship between them, called hierarchical ranking. To the best of our knowledge, this is the first work to address this task with the three levels jointly. 
    
    \item To implement the hierarchical ranking strategy, we propose three schemes: the scheme under Multi-Task Learning (MTL), Ranking Integration (RI) and Progressive Ranking Integration (PRI).

    \item Extensive experiments are conducted on public datasets, TREC-QA and WikiQA. Our model achieves state-of-the-art performance on both WikiQA and TREC-QA. Results demonstrate the effectiveness of the proposed hierarchical ranking. 
\end{itemize}

The rest of the paper is organized as follows: we introduce related work about answer selection task in section 2. In section 3, we introduce the proposed strategy in detail. Then the experimental analysis and comparisons of the methods are reported in section 4. At last, we draw a conclusion for the entire paper.

\section{Related Work}

\noindent\textbf{Mainstream Architectures.} The Siamese Network \cite{bromley1993signature} is a popular architecture. It calculates the relevance score, using such as Euclidean distance, cosine similarity, etc, according to the sentence vectors of the given question and answer. Many works are based on Siamese Network \cite{yin2015abcnn:,santos2016attentive,rao2016noise-contrastive,tan2015lstm-based}.

A disadvantage of siamese architecture is that some key information may be lost after feature extraction of sentences to obtain sentence vectors. To solve this problem, another architecture has been proposed, called Compare-Aggregate Network \cite{wang2016a}. In the Compare-Aggregate Network, matching is based on smaller units such as words or phrases and the comparison process is more specific. Thus, the compare-aggregate network performs better. Many recent work is based on this architecture\cite{wang2016a,wang2017bilateral,bian2017a,yoon2019a}. Considering the flexible structure and excellent performance of this architecture, we build models based on this architecture to validate our ideas.

\noindent\textbf{Mainstream Approaches.} Previous work typically addressed the task of answer selection by three approaches: pointwise, pairwise and listwise. Benefiting from the development of neural networks, all three approaches have achieved good performance.

The pointwise approach \cite{devlin2018bert:,tay2017learning,wang2017bilateral,shen2017inter-weighted,deng2019multi-task,shen2018knowledge-aware} usually trains a binary classification model or a logical regression model by cross-entropy loss. This approach regards the question and its candidate answers as separate training pairs, which simplifies the training and is easy to utilize many other classification models. 

The pairwise approach \cite{rao2016noise-contrastive,he2015multi-perspective,santos2016attentive} learns the correlation among candidates in a contrastive way. Compared with the pointwise method, this approach employs the supervisory signals by connecting the positive answers and negative answers in the candidates. 

The listwise approach \cite{bian2017a,wang2016a} aims to learn the group structure information of the question-answer pairs. This approach could compare all the candidate answers globally. For more details, we refer interested readers to \cite{LiLearning,liu2009learning}.

\section{Model Architecture}

In this section, we first formulate the task, then a brief description of the backbone network is given. Next, We will describe the scheme under multi-task learning strategy, ranking integration scheme and progressive ranking integration scheme, and explain their differences and the motivations behind them.

\subsection{Formulation}

For a given question and a candidate answer, let $Q$ denote the question, and $A$ denote the answer, $y$ is the ground-truth label representing whether $A$ is positive or not. 

\subsection{Backbone Network}

Considering the excellent performance of the compare-aggregate model \cite{wang2016a,bian2017a,yoon2019a}, we take it as the backbone and prove the effectiveness of the proposed hierarchical ranking strategy. A typical compare-aggregate model has the following layers: encoding layer, interaction layer, comparison and aggregation layer and prediction layer.

\subsubsection{Encoding Layer} 
In this layer, for the given question $Q=\{q_1,q_2,...,q_n\}$ and and an answer $A=\{a_1,a_2,...,a_m\}$. Firstly we map them into word embedding sequences $E^q=\{\mathbf{e^q_1},\mathbf{e^q_2},...,\mathbf{e^q_n}\}$ and $E^a=\{\mathbf{e^a_1},\mathbf{e^a_2},...,\mathbf{e^a_m}\}$ for $Q$ and $A$ respectively. Then, we use a modified version of LSTM/GRU \cite{wang2016a} to obtain the context representation $H^q$ and $H^a$ which contains context information: 

\begin{equation}
\begin{split}
    H^a &= sigmoid(E^aW_1+b_1) \odot tanh(E^aW_2+b_2)\\
    H^q &= sigmoid(E^qW_1+b_1) \odot tanh(E^qW_2+b_2)
\end{split}
\end{equation}
where $W_1$, $W_2$, $b_1$ and $b_2$ are parameters in the modified LSTM/GRU.

\subsection{Interaction Layer}
After obtaining the representation of the input question and answer, we utilize a standard co-attention module to learn the interactions between question-answer pair:
\begin{equation}
%\begin{center}
\begin{split}
    M = {H^q}{H^a}^\mathrm{T}
\end{split}
%\end{center}
\end{equation}

Then a soft-align is used to obtain the alignment $\hat H^q$ and $\hat H^a$ to each other as follows:
\begin{equation}
%\begin{center}
\begin{split}
    \hat{H}^q &= softmax(M)H^a\\
    \hat{H}^a &= softmax(M^{T})H^q
\end{split}
%\end{center}
\end{equation}

\subsubsection{Comparison and Aggregation Layer}\label{calayer}

To compare the aligned question $\hat{H}^q$ and $H^q$ on word-level, we adopt a comparison function to get the matching sequences. So does the aligned answer $\hat{H}^a$ and $H^a$.
\begin{equation}
%\begin{center}
\begin{split}
    C^q &= \hat{H}^q \odot  H^q \\
    C^a &= \hat{H}^a \odot  H^a 
\end{split}
%\end{center}
\end{equation}
where the operator $\odot$ is element-wise multiplication. $C^{a}=\{\mathbf{c_1^a},\mathbf{c_2^a},...,\mathbf{c_m^a}\}$ and $C^{q}=\{\mathbf{c_1^q},\mathbf{c_2^q},...,\mathbf{c_n^q}\}$ are feature sequences. Then,we apply a one-layer CNN to aggregate the matching sequence to a fixed dimensional vector.

\begin{equation}
%\begin{center}
\begin{split}
   \mathbf{r^{q}}&=CNN\left ( [ \mathbf{c_{1}^{q}},\mathbf{c_{2}^{q}},...,\mathbf{c_{n}^{q}}\right ])\\
   \mathbf{r^{a}}&=CNN\left ( [ \mathbf{c_{1}^{a}},\mathbf{c_{2}^{a}},...,\mathbf{c_{m}^{a}} \right ])
\end{split}
%\end{center}
\end{equation}

We concatenate $\mathbf{r^q}$ and $\mathbf{r^a}$ together as a matching vector $\mathbf{r}=[\mathbf{r^q};\mathbf{r^a}]$. 

\subsubsection{Prediction Layer}

In the prediction layer, we pass $\mathbf{r}$ through a multi-layer perceptron for prediction.

\begin{figure}[t]
\centering
\includegraphics[width=0.45\textwidth]{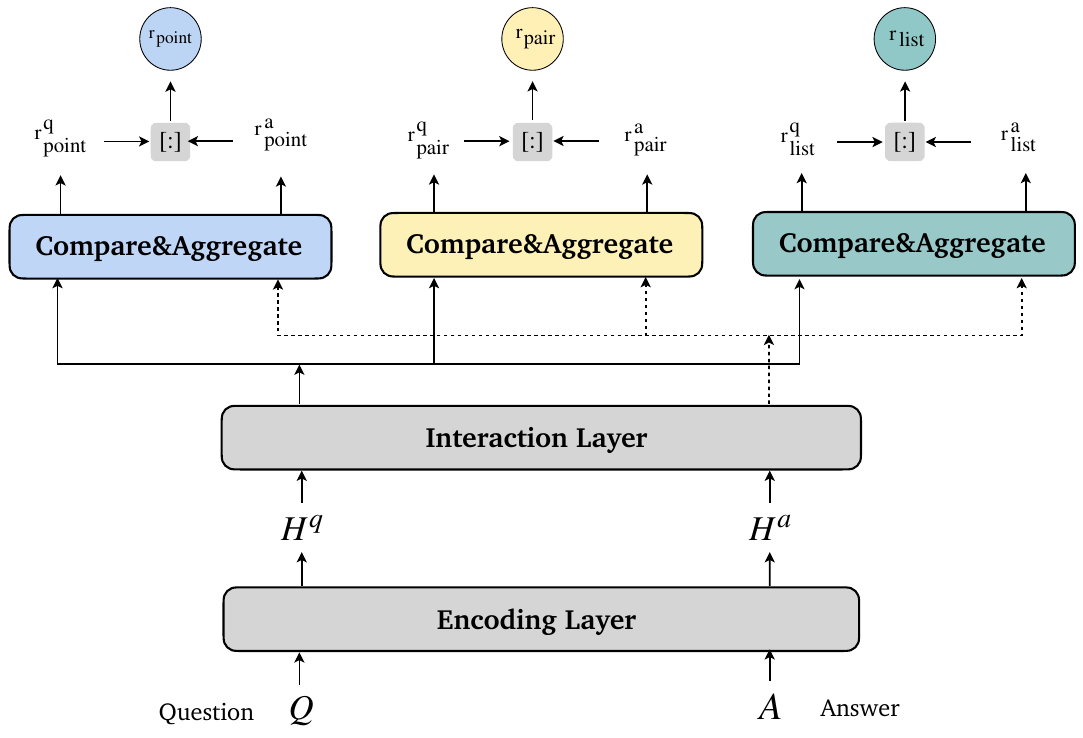}
\caption{The architecture of the model that follow MTL strategy. $[;]$ denotes the vector concatenate operation.}
\label{network_MTL}
\end{figure}

\subsection{Multi-task Learning Strategy}

As mentioned before, multi-task learning (MTL) is an intuitive solution of using the three levels of ranking jointly. The architecture of the model which follows MTL is shown in Figure \ref{network_MTL}. It follows hard parameter sharing. In MTL Strategy, the three levels of ranking are treated as three tasks.

Considering that rankings on different granularities have different concerns, the encoding layer and the interaction layer are shared between the three rankings, and the comparison and aggregation layers and prediction layers are level-specific. Specifically, the advantage of sharing the encoding layer and the interaction layer is that the low-level representation could benefit from all the three ranking objectives. The goal of the comparison and aggregation layer is to obtain the comparison features of the question and answer, and the comparison features based on different granularity are distinct. Therefore, applying the level-specific comparison and aggregation layer is helpful to learn specific features from different perspectives.

With the three level-specific comparison and aggregation layers, we can obtain three comparison features. Let $\mathbf{r_{point}}$ denote the feature for point-level ranking, $\mathbf{r_{pair}}$ denote the feature for pair-level ranking and $\mathbf{r_{list}}$ as the feature used for list-level ranking. Note that the three comparison and aggregation layers for the different rankings have the same architecture. Next, we describe the level-specific prediction layer and the losses used for training. At each time, we feed a given question and all its candidate answers to the model, thus, we suppose there are $k$ positive answers and $t$ negative answers in the candidate list for clearer description. 

\subsubsection{Point-level Ranking}\label{point-prediction} The point-level ranking strategy aims to optimize the predicted probability distribution of each question-answer pair to be consistent with the ground truth. Thus, we regard the task as a classification problem. For the $i$-th question-answer pair, We feed $\mathbf{r^i_{point}}$ into a two-layer perceptron for prediction and employ cross-entropy loss for training.

\begin{equation}
\begin{split}
   \mathbf{p^i_{point}} &= softmax(MLP(\mathbf{r^i_{point}})) \\
   L_{point} &= -\frac{1}{k+t}\sum^{k+t}_{i}logP(\mathbf{y_{i}}|\mathbf{p^i_{point}})
   \label{pointout}
\end{split}
\end{equation}
where $\mathbf{y_{i}}$ is the ground-truth label for the $i$-th question-answer pair. In the testing phase, the scores used for sorting candidate answers are the probabilities that the answer is positive.

\begin{figure*}[t]
\centering
\includegraphics[width=0.8\textwidth]{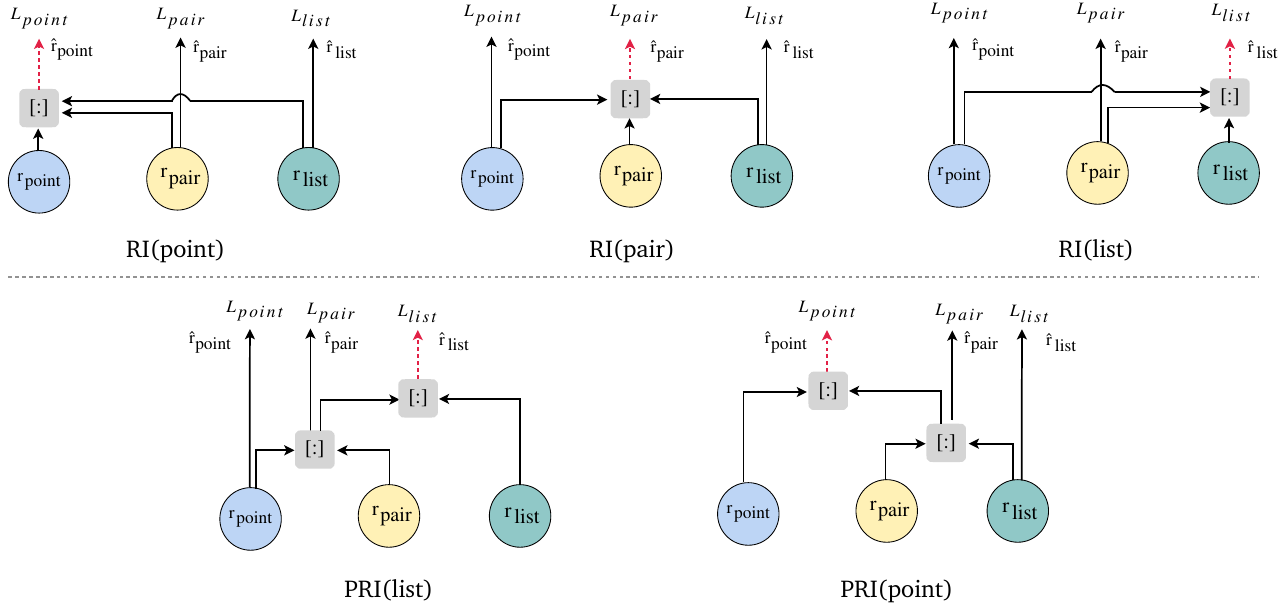}
\caption{The proposed three RI schemes and two PRI schemes. The red dashed line represents the integrated representation employed in testing. Note that the architecture of the bottom layers in RI and PRI is the same as the MTL model so it’s omitted for conciseness.}
\label{network_3}
\end{figure*}

\subsubsection{Pair-level Ranking}\label{pair-prediction} The pair-level ranking strategy is to optimize the relative ordering of a pair of prediction scores. The main idea behind it is noise contrastive estimation. For the $i$-th question-answer pair, we feed $\mathbf{r^i_{pair}}$ into a two-layer perceptron to calculate a relevance score.

\begin{equation}
   \mathbf{p^i_{pair}} = MLP(\mathbf{r^i_{pair}} )
\end{equation}

Here, we introduce two methods to construct the pairs for pair-level ranking. In the first method, we generate all possible positive-negative pairs. In the second method, we pick the negative pair with the highest relevance score and assign scores of all positive pairs higher than it. For the given $k$ positive answers and $t$ negative answers, we can obtain $k*t$ pairs in the first method and $k$ pairs in the second method.

For a positive-negative pairs, let $\mathbf{p_{pair}^{+}}$ denote the score of the positive pair and $\mathbf{p_{pair}^{-}}$ denote the relevance score of the negative pair. We use a margin loss to assign larger scores to positive pairs than negative pairs for each $(\mathbf{p_{pair}^{+}},\mathbf{p_{pair}^{-}})$. The loss for pair-level ranking can be formulated as follow.

\begin{equation}
\begin{split}
    L_{pair} &= \frac{1}{k\ast t} \sum^{k}\sum^{t} max(0,M-(\mathbf{p_{pair}^{+}}-\mathbf{p_{pair}^{-}}))
    \label{pairout}
\end{split}
\end{equation}
where $M$ refers to the margin which treated as a hyper-parameter. Note that the above loss follows the first method of positive-negative pairs generation. In the testing phase, $\mathbf{p^i_{pair}}$ is used for sorting candidate answers.

\subsubsection{List-level Ranking.}\label{list-prediction} The list-level ranking strategy is to optimize the relative ordering of the list of prediction scores. We feed $\mathbf{r^i_{list}}$ into a two-layer perceptron to calculate a relevance score. Then We calculate the probability of $k+t$ pairs to be the positive one and utilize it to rank candidate answers. We utilize KL-divergence loss for training. Besides, Label $Y$ needs to be normalized.

\begin{equation}
\begin{split}
   \bm{ p^i_{list}} &= MLP(\mathbf{r^i_{list}})\\
   p_{list} &= softmax(\left\{\bm{ p^1_{list}},...,\bm{ p^{k+t}_{list}}\right\})\\
   Y &= \frac{Y}{\sum_{i=1}^{k+t}y_{i}} \\
    L_{list} &= \frac{1}{k+t}KL(p_{list}||Y) \label{listout}
\end{split}
\end{equation}   

In the testing phase, $p_{list}$ is the list of scores used for sorting candidate answers.

\subsubsection{Joint Loss.} The MTL strategy is to minimize the joint loss.
\begin{equation}
\begin{split}
    L &= \lambda_{point}{L_{point}}+\lambda_{pair}{L_{pair}}+\lambda_{list}{L_{list}}
\end{split}
\end{equation}
where $\lambda_{point},\lambda_{pair},\lambda_{list}$ balance the effect of each loss.

\subsubsection{An Example about Batch Working.}

Here, we give an example to clarify how batching works for the different ranking strategy. Suppose a given question has 2 positive answers and 3 negative answers. There are 5 question-answer pairs for point-level ranking and one list of question-answer pairs for list-level ranking. For pair-level ranking, there are 6 or 3 positive-negative pairs for training, depending on the method of positive-negative pairs generation. For the convenience of explanation, assume that each question in the dataset has 2 positive answers and 3 negative answers, then, if the batch size is set to 10, for point-wise approach, there are 50 question-answer pairs; for pair-wise approach, there are 60 or 30 positive-negative pairs and 10 lists of pairs for list-wise approach. As we can see from this example, the amount of data used for different ranking training is different on different granularities.

\subsection{Ranking Integration}
As mentioned before, the three different ranking strategies can complement and reinforce each other to achieve the goal of ranking the list of candidate answers. The multi-task learning strategy is insufficient to make full use of these internal relationships. In this section, we explore several more direct schemes, called Ranking Integration (RI). 

An intuitive experience is, the extracted features are different under different optimization objectives. Thus, the three levels of optimization goals can learn features from different perspectives. The features obtained by point-level methods contain information about the relevance between $Q$ and $A$. The pair-level features contain the comparison information between an opposite pair $(Q,A^{+})$ and $(Q,A^{-})$. The list-level features contain information about comparing $(Q,A)$ with all other pairs. Based on these considerations, we argue that it is reasonable to integrate the features from the three approaches. The enhanced feature will be helpful to improve the performance of the model. This is where the idea of RI scheme comes from.

As shown in the top row of Figure \ref{network_3}, RI contains three schemes. In each RI scheme, we take one level ranking as the main objective and the other two rankings as the auxiliary objectives. The main objective is used for the final prediction, and the auxiliary objectives are used for training the distinct matching features on the specific granularities and providing more evidence for the main objective. The main difference between RI and MTL is that the matching feature used for the final prediction is enhanced in RI. Next, we will describe the feature enhancement scheme in RI(point), RI(pair) and RI(list).

In RI(point), we take the point-level ranking as the main objective. The features extracted from the pair-level ranking and list-level ranking are used to enhance the feature used for point-level ranking.
\begin{equation}
\begin{split}
    \mathbf{\hat{r}_{pair}}&=\mathbf{r_{pair}}\\ 
    \mathbf{\hat{r}_{list}}&=\mathbf{r_{list}}\\ \mathbf{\hat{r}_{point}}&=[\mathbf{\hat{r}_{pair}};\mathbf{\hat{r}_{list}};\mathbf{r_{point}}]
\end{split}
\end{equation}
where $\mathbf{\hat{r}_{point}}$ is the enhanced feature. We pass $\mathbf{\hat{r}_{pair}}$ and $\mathbf{\hat{r}_{list}}$ through the level-specific prediction layer for the training of auxiliary objectives, and pass $\mathbf{\hat{r}_{point}}$ through the point-level prediction layer for the final prediction. The losses used to train different rankings are the same as those described in MTL strategy.

Similarly, in RI(pair), we take the pair-level ranking as the main objective.
\begin{equation}
\begin{split}
    \mathbf{\hat{r}_{point}}&=\mathbf{r_{point}}\\ \mathbf{\hat{r}_{list}}&=\mathbf{r_{list}}\\ \mathbf{\hat{r}_{pair}}&=[\mathbf{\hat{r}_{point}};\mathbf{\hat{r}_{list}};\mathbf{r_{pair}}]
\end{split}
\end{equation}

In RI(list), we take the list-level ranking as the main objective.
\begin{equation}
\begin{split}
    \mathbf{\hat{r}_{point}}&=\mathbf{r_{point}}\\ \mathbf{\hat{r}_{pair}}&=\mathbf{r_{pair}}\\ \mathbf{\hat{r}_{list}}&=[\mathbf{\hat{r}_{point}};\mathbf{\hat{r}_{pair}};\mathbf{r_{list}}]
\end{split}
\end{equation}

\subsection{Progressive Ranking Integration}

In this section, we introduce the proposed novel Progressive Ranking Integration (PRI). As shown in the bottom row of Figure \ref{network_3}, PRI contains two schemes. Compared with RI, PRI employs a progressive way to integrate features and compute losses. The idea is borrowed from the divide-and-conquer technique. For a given question and its candidate answers, the point-level ranking strategy concentrates on determining the relevance between the question and an answer. We can get the ranking results of a positive pair and a negative pair (pair-level ranking) by combining the results of relevance judgment in pairs. In other words, the solution of the point-level ranking strategy can be combined to give a solution to the pair-level ranking. Similarly, combining the ranking results of multiple positive-negative pairs can get the ranking results of a list of candidate answers. In other words, the solution of the pair-level ranking strategy can be combined to give a solution to the list-level ranking. This follows the divide-and-conquer technique. We implement this idea in PRI(list). Besides, we also explore an inverse process of PRI(list), called PRI(point). The intuition behind PRI(point) is that it is also reasonable to sort the candidate answers altogether (list-level) firstly, then gradually refine it to sort the pairs, and finally optimize the similarity of each question-answer pair. In conclusion, PRI(list) aims to rank the candidate answers from individual to list and PRI(point) aims to obtain the ranking results by progressive refinement. 

\subsubsection{PRI(list)}
Follow the idea of PRI(list), fine-grained ranking can be treated as the foundation for the next level ranking. More formally:

\begin{equation}
\begin{split}
   \mathbf{\hat{r}_{point}} &= \mathbf{r_{point}}\\
   \mathbf{\hat{r}_{pair}} &= [\mathbf{\hat{r}_{point}};\mathbf{r_{pair}}]\\
   \mathbf{\hat{r}_{list}} &= [\mathbf{\hat{r}_{pair}};\mathbf{r_{list}}]
\end{split}
\end{equation}

\noindent\textbf{PRI(point).} In this scheme, we explore an inverse process of PRI(list) as follow, and take the output of point-level ranking as the final prediction.

\begin{equation}
\begin{split}
   \mathbf{\hat r_{list}} &= \mathbf{r_{list}}\\
   \mathbf{\hat r_{pair}} &= [\mathbf{\hat{r}_{list}};\mathbf{r_{pair}}]\\
   \mathbf{\hat r_{point}} &= [\mathbf{\hat{r}_{pair}};\mathbf{r_{point}}]
\end{split}
\end{equation}

\begin{table*}[t!]
\begin{center}
\begin{tabular} {|l|cc|cc|}
\hline
     \multirow{2}{*}{Models} &
    \multicolumn{2}{c|}{WikiQA} & \multicolumn{2}{c|}{TREC-QA}\\
    %\midrule{2-13}
    & MAP & MRR & MAP & MRR  \\
    \hline
    AP-CNN\cite{santos2016attentive} & 0.689 & 0.696 &0.753  & 0.851\\
    MP-CNN\cite{he2015multi-perspective} & 0.693 & 0.709 & 0.777 &0.836\\
    NCE\cite{rao2016noise-contrastive} & 0.701  & 0.718 & 0.801  & 0.877\\
    L.D.C\cite{wang2016sentence} & 0.706 & 0.723  & 0.771  & 0.845\\
    PWIM\cite{he2016pairwise} & 0.709 & 0.723 & - & - \\
    HyperQA\cite{tay2018hyperbolic} & 0.712  & 0.727  & 0.784  & 0.865\\
    BIMPM\cite{wang2017bilateral} &  0.718  & 0.731  & 0.802  & 0.875\\
    IWAN\cite{shen2017inter-weighted} & 0.733  & 0.750  & 0.822  & 0.889\\ 
    DCA\cite{bian2017a} & 0.736  & 0.749   & 0.813  & 0.867 \\
    IWAN+sCARNN\cite{tran2018the} &-&-&0.829&0.875\\
    MCAN\cite{tay2018multi-cast} &-&-&0.838&0.904\\
    \hline
    BERT Fine-Tuning\cite{2020Contextualized} & 0.843 & 0.857 & 0.905 & 0.967 \\
    RoBERTa Fine-Tuning\cite{2020Contextualized} & 0.900 & 0.915 & 0.936 & 0.978 \\
    \hline
    CA(point)& 0.721 & 0.735 & 0.809 & 0.867\\
    CA(pair) & 0.725 & 0.737 & 0.832 & 0.889\\
    CA(list) & 0.736 & 0.749 & 0.809 & 0.863 \\
    \hline
    MTL(point)& 0.734 & 0.747 & 0.831 & 0.885 \\
    MTL(pair)& 0.734 & 0.745 & 0.837 & 0.895 \\
    MTL(list)& 0.739 & 0.751 & 0.835 & 0.892 \\
    \hline
    RI(point) & 0.734 & 0.747 & 0.823 & 0.880 \\
    RI(pair) & 0.737 & 0.749 & \bf{0.842} & \bf{0.904} \\
    RI(list) & 0.740 & 0.751  & 0.835 & 0.892 \\
    
    \hline
    PRI(point) & 0.732 & 0.743 & 0.825 & 0.890 \\
    PRI(list) & \bf{0.742}$^\dag$ & \bf{0.754}  & 0.841$^{\ddag}$ & 0.898$^{\ddag}$ \\
    \hline
\end{tabular}
\end{center}
\caption{Evaluation results on WikiQA and Clean version TREC-QA in terms of MAP and MRR. We conduct T-test comparing PRI(list) and a strong baseline DCA. The marker $^{\dag}$ means p-value $p<0.1$ and $^{\ddag}$ means p-value $p<0.05$.}
\label{table_total}
\end{table*}

\section{Experiments}

In this section, we first introduce the datasets and evaluation metrics used in the experiment. Then, the model parameters, training settings are introduced in detail. Finally, we analyze the experimental results and validate our methods.

\subsection{Datasets and Metrics} %改完

We utilize two public datasets, TREC-QA and WikiQA. Many previous works also used these two datasets to evaluate model performance.

The TREC-QA dataset \cite{wang2007what}, collected from TREC-QA track 8-13 data. In this paper, we only display the results on the Clean version TREC-QA which consists of 1,229 questions with 53,417 question-answer pairs in train set, 65 questions with 1,117 pairs in development set and 68 questions with 1,442 pairs in the test set.

WikiQA \cite{yang2015wikiqa:}, a common dataset for answer selection.  We follow \cite{yang2015wikiqa:} to remove all questions with no correct candidate answers. The excluded WikiQA has 873 questions with 8627 question-answer pairs in the train set, 126 questions with 1130 pairs in the development set and 243 questions with 2351 pairs in the test set. 

Follow previous works, MAP(mean average precision) and MRR(mean reciprocal rank) are adopted as our evaluation metrics.

\subsection{Compared Methods}

We compare the proposed scheme against competitive baselines follow \footnote{https://aclweb.org/aclwiki/Question\_Answering\_(State\_of\_the\_art)}. It should be noted that some other research lines utilize transfer learning \cite{lai-etal-2019-gated}, external knowledge \cite{madabushi2018integrating,shen2018knowledge-aware} such as knowledge graphs or pre-trained model BERT \cite{lai-etal-2019-gated,shao-etal-2019-aggregating}, which are good works. But for fairness, we do not use these models as baselines. Besides, for a fair comparison, we reimplement DCA \cite{bian2017a} but not clipping the number of candidate answers corresponding to each question to the same number. We also present
the current state-of-the-art BERT \cite{devlin2018bert:} fine-tuning results and RoBERTa \cite{liu2019roberta:} fine-tuning results, which reported in \citeauthor{2020Contextualized}, on each dataset.

\subsection{Our Methods}
To verify the strategies we proposed, we implement several models as follow. Note that the level indicated in parentheses is the ranking level used for the final prediction.
\begin{itemize}
    \item \textbf{CA.} CA(point), CA(pair) and CA(list) are the basic compare-aggregate models which follow only one specific ranking for training respectively.
    \item \textbf{MTL.} MTL(point), MTL(pair) and MTL(list) are the models which follow the MTL strategy. 
    \item \textbf{RI.} RI(point), RI(pair) and RI(list) are the models which follow the RI scheme.
    \item \textbf{PRI.} PRI(point) and PRI(list) are the models which follow the PRI scheme.
\end{itemize}

\subsection{Implementation details}

We implement our model with Pytorch(version 1.5.0) and train on one Tesla P100 GPU. We follow \citeauthor{bian2017a} to tokenize and pad the sentences, but not clipping the number of candidate answers corresponding to each question to the same number.

\noindent\textbf{Model details.} For both WikiQA and TREC-QA, the pre-trained 300-dimensional GloVe word vectors \cite{pennington2014glove:} on the 840B Common Crawl corpus are used as initialization for word embedding and embeddings of out-of-vocabulary words are initialized to zeros. For TREC-QA, all embeddings are fixed during training. For WikiQA, all embeddings are fine-tuned. The dimension of hidden states is 300. We use [1,2,3,4,5] as the kernel size of the one-layer CNN. The output channel of CNN is 150. 
For WikiQA, we adopt k-max \cite{bian2017a} to filter irrelevant words. In the pair-level ranking, for WikiQA, we apply the margin loss for each positive-negative pair and the prediction scores are normalized with $sigmoid$. The margin is set to 0.8. For TREC-QA, we adopt the optional solution which picks the negative case with the highest relevance score and assigns all positive individual scores higher than it. The margin is set to 1.

\begin{table}[t!]
\begin{center}
\begin{tabular} {|l|cc|cc|}
\hline
    \multirow{2}{*}{Models} &
    \multicolumn{2}{c|}{WikiQA} & \multicolumn{2}{c|}{TREC-QA}\\
    & MAP & MRR & MAP & MRR  \\

    \hline
    PRI(list) & 0.742 & 0.754  & 0.841 & 0.898 \\
    \hline
    (1) w/o point-level & 0.735 & 0.746 & 0.829 & 0.883 \\
    (2) w/o pair-level & 0.739 & 0.752 & 0.833 & 0.881 \\
    (3) PRI(all list) & 0.726 & 0.737  & 0.814 & 0.859 \\

    \hline
\end{tabular}
\end{center}
\caption{Ablation study.}
\label{table_ablation}
\end{table}

\noindent\textbf{Training details.}
For training, we adopt Adam to optimize the models. The learning rate for model parameters is 5e-4. For wikiQA, the learning rate for embeddings is 5e-5. For TREC-QA, embeddings are not updated during training. Mini-batch is taken to train the models, each batch contains 30 questions and their candidate answers. We use an early stop to prevent overfitting. We set the early stop to 10, which means we stop training when the MAP of development set stops increasing for 10 epochs in succession. We set $\lambda_{pair}$ to 1, $\lambda_{list}$ to 1. For WikiQA, we set $\lambda_{point}$ to 2, and 1 for TREC-QA. All the results we reported are the average of 5 runs with random seeds [0,1,2,3,4].

\subsection{Experimental Analysis}
The experimental results are reported in Table \ref{table_total} and \ref{table_ablation}. The best scores (besides BERT) on each metric are marked in bold. All the results are the average score of 5 runs.

\noindent\textbf{Compare with baselines.} 
Table \ref{table_total} illustrates the results compared with baselines. Firstly, we observe that it is effective to use the three levels of ranking jointly for training. In this paper, we propose the scheme under MTL strategy, RI scheme and PRI scheme, and all of them outperform the CA models which follow only one specific ranking. Specifically, the performances of MTL(list), RI(list) and PRI(list) are better than that of CA(list) on both WikiQA and TREC-QA in terms of MAP and MRR, and similarly, the performances of MTL(point), RI(point) and PRI(point) are also better than CA(point). This proves that the main ranking objectives can benefit from the other two auxiliary objectives. As such, the three schemes we proposed for hierarchical ranking are effective and crucial. Secondly, we observe that in some cases, RI performs better than MTL. RI can improve performance marginally. However, it also contributes to performance. Thirdly, we can observe that PRI can bring further performance improvement based on RI. Especially, compared RI(list), PRI(list) performs better on both two datasets. Finally, we observe that except for BERT, the proposed PRI(list) achieves the best performance on all metrics on WikiQA and RI(pair) achieves the best performance on all metrics on TREC-QA. Besides, PRI(list) achieves the second-best performance on MAP on TREC-QA and the performance of RI(list) is also better than other baselines (besides BERT). Specifically, compared with a strong baseline DCA on TREC-QA, our method PRI(list) significantly outperforms it by +0.028 on MAP and +0.031 on MRR. This shows that our network which employs the hierarchical ranking strategy is successful. 

\noindent\textbf{Ablation study.} In this section, we aim to demonstrate the effectiveness of the joint ranking. We take the best performing PRI(list) as the full configuration model, and the following models are constructed: (1) ``w/o point-level'': we remove the point-level ranking part in PRI(list) and keep only the pair-level ranking as an auxiliary objective. (2) ``w/o pair-level'': we remove the pair-level ranking part in PRI(list) and keep only the point-level ranking as an auxiliary objective. (3) ``PRI(all list)'': We set all the three objectives as list-level ranking. The results are reported in Table \ref{table_ablation}.

Firstly, we observe that both the removal of point-level ranking or pair-level ranking caused performance degradation. This shows that both the point-level ranking and pair-level ranking are necessary.

Secondly, we observe that replacing all three rankings with list-level ranking significantly decreases the performance. This proves that it is necessary to apply the three levels of ranking jointly in PRI scheme, and the performance improvement comes from the hierarchical ranking strategy rather than more parameters.

\begin{figure}[t]
\centering
\includegraphics[width=0.4\textwidth]{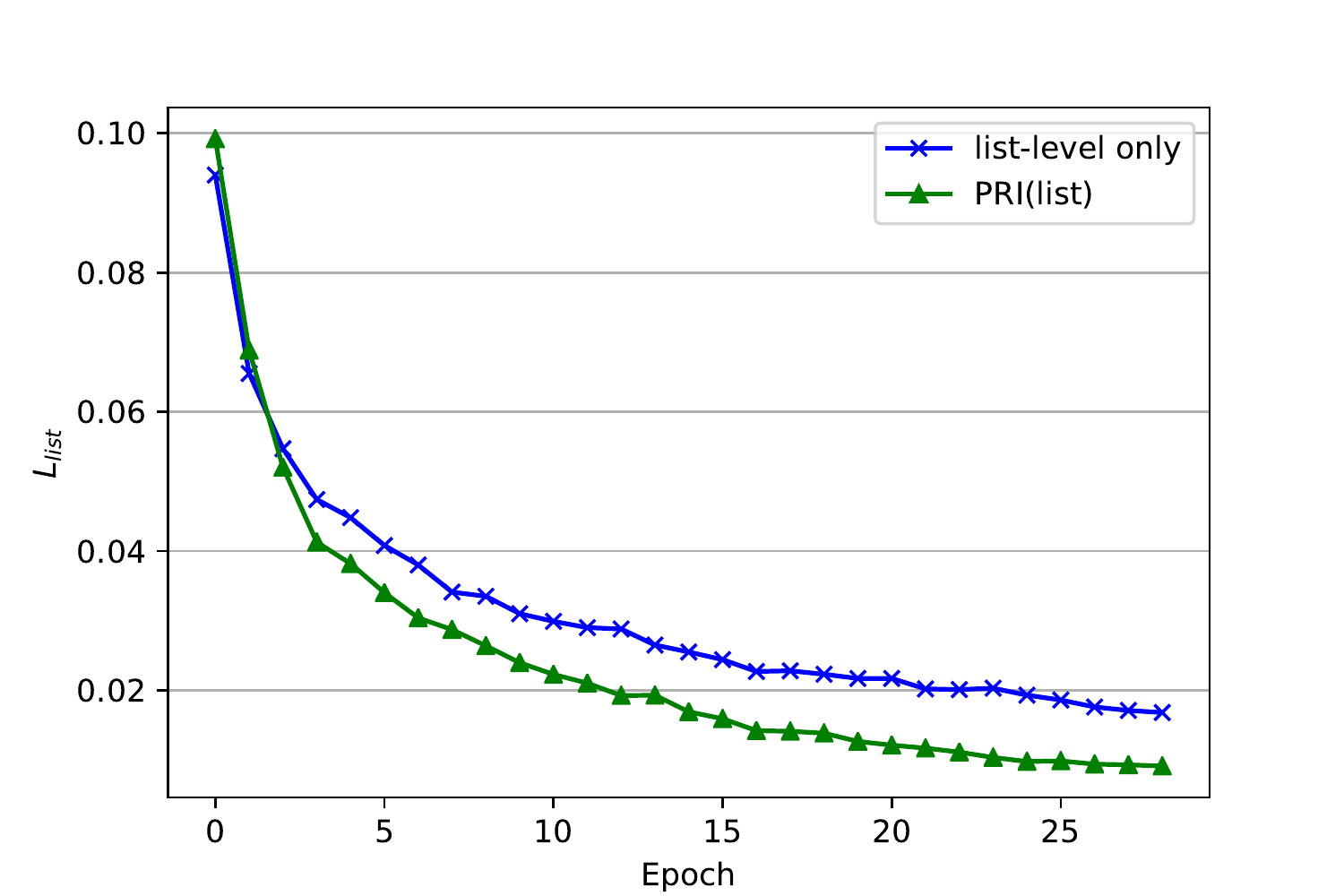}
\caption{$L_{list}$ loss curves in different strategy.}
\label{figure_loss}
\end{figure}

\noindent\textbf{Loss.} As shown in Figure \ref{figure_loss}, we plot the curves of $L_{list}$ of different schemes. We observe that $L_{list}$ in PRI(list) scheme decreases faster than it in ``list-level only''. This shows that with the help of multiple approaches and feature integration, our method PRI(list) needs fewer epochs to converge.

\section{Conclusion}
In this paper, we propose a novel strategy for answer selection, called hierarchical ranking. There are three levels of ranking in the proposed strategy: point-level ranking, pair-level ranking, and list-level ranking. To implement the hierarchical ranking strategy, we first introduce a scheme under Multi-Task Learning (MTL) strategy, then propose Ranking Integration (RI) scheme, and furthermore, we explore the idea of integrating the feature progressively via Progressive Ranking Integration (PRI). Experimental results demonstrate that the proposed hierarchical ranking strategy is effective, all the three schemes under hierarchical ranking strategy outperform the models which follow only one specific ranking approach, and the proposed RI and PRI can further improve the performance. Our method achieves state-of-the-art (non-BERT) performance on both TREC-QA and WikiQA.

\end{document}